\documentclass[sigconf,natbib=true]{acmart}

\usepackage{multirow}
\usepackage[most]{tcolorbox}
\usepackage{rotating}
\usepackage{xspace}
\usepackage{placeins}
\usepackage{xcolor}

\usepackage{tabularx}
\tcbset{
  mylisting/.style={
    colback=blue!10,
    colframe=blue!50!black,
    sharp corners,
    boxrule=0.5pt,
    enhanced,
    listing only,
    fontupper=\ttfamily\small,
  }
}

\tcbset{
  codelisting/.style={
    colback=gray!10,
    colframe=gray!50!black,
    sharp corners,
    boxrule=0.5pt,
    enhanced,
    listing only,
    fontupper=\ttfamily\small,
  }
}

\AtBeginDocument{%
  }

\copyrightyear{2026}
\acmYear{2026}
\setcopyright{cc}
\setcctype{by}
\acmConference[ICTIR '26]{Proceedings of the 2026 International ACM SIGIR Conference on Innovative Concepts and Theories in Information Retrieval (ICTIR)}{July 25, 2026}{Melbourne, VIC, Australia}
\acmBooktitle{Proceedings of the 2026 International ACM SIGIR Conference on Innovative Concepts and Theories in Information Retrieval (ICTIR) (ICTIR '26), July 25, 2026, Melbourne, VIC, Australia}
\acmDOI{10.1145/3805713.3820425}
\acmISBN{979-8-4007-2600-2/2026/07}

\begin{document}

\newcommand{\approach}{\textsc{\textcolor{RawSienna}{DoGMaTiQ}}\xspace}
\newcommand{\commona}{\textsc{\textcolor{RawSienna}{Common}}\xspace}
\newcommand{\randoma}{\textsc{\textcolor{RawSienna}{Sample}}\xspace}
\newcommand{\approachl}{\textsc{\textcolor{RawSienna}{DoGMaTiQL}}\xspace}
\newcommand{\commonl}{\textsc{\textcolor{RawSienna}{CommonL}}\xspace}
\newcommand{\randoml}{\textsc{\textcolor{RawSienna}{SampleL}}\xspace}
\newcommand{\argue}{\texttt{argue-eval}\xspace}
\newcommand{\autoargue}{\textsc{Auto}-ARGUE\xspace}
\newcommand{\crucible}{\textsc{Crucible}\xspace}

\title{\approach: Automated Generation of Question-and-Answer Nuggets for Report Evaluation}

\author{Bryan Li}
\authornote{Work done while at the University of Pennsylvania, Email: \texttt{bryanli.ca@gmail.com}} 
\orcid{0000-0002-5779-1662}
\affiliation{%
  \institution{Google Inc.}
  \city{New York}
  \state{NY}
  \country{USA}
}

\author{William Walden} 
\authornote{Corresponding author. Email: wwalden1@jh.edu}
\affiliation{%
  \institution{Johns Hopkins University}
  \city{Baltimore}
  \state{MD}
  \country{USA}}

\author{Yu Hou}
\affiliation{%
  \institution{University of Maryland}
  \city{College Park}
  \state{MD}
  \country{USA}}

\author{Gabrielle Kaili-May Liu}
\affiliation{%
  \institution{Yale University}
  \city{New Haven}
  \state{CT}
  \country{USA}}

\author{Dawn Lawrie}
\affiliation{%
  \institution{Johns Hopkins University}
  \city{Baltimore}
  \state{MD}
  \country{USA}}

\author{James Mayfield}
\affiliation{%
  \institution{Johns Hopkins University}
  \city{Baltimore}
  \state{MD}
  \country{USA}}

\author{Eugene Yang}
\affiliation{%
  \institution{Johns Hopkins University}
  \city{Baltimore}
  \state{MD}
  \country{USA}}

\author{Chris Callison-Burch}
\affiliation{%
  \institution{University of Pennsylvania}
  \city{Philadelphia}
  \state{PA}
  \country{USA}}

\author{Laura Dietz}
\affiliation{%
  \institution{University of New Hampshire}
  \city{Durham}
  \state{NH}
  \country{USA}}

\renewcommand{\shortauthors}{Bryan Li et al.}

\begin{abstract}
Evaluation of long-form, citation-backed reports has lately received significant attention due to the wide-scale adoption of retrieval-augmented generation (RAG) systems. Core to many evaluation frameworks is the use of atomic facts, or \emph{nuggets}, to assess a report's coverage of query-relevant information attested in the underlying collection. While nuggets have traditionally been represented as short statements, recent work has used question-answer (QA) representations, enabling fine-grained evaluations that decouple the information need (i.e.\ the question) from the potentially diverse content that satisfies it (i.e.\ its answers).

A persistent challenge for nugget-based evaluation is the need to manually curate \emph{sets} of nuggets for each topic in a test collection---a laborious process that scales poorly to novel information needs. This challenge is  acute in cross-lingual settings, where information is found in multilingual source documents. Accordingly, we introduce \approach, a pipeline for generating high-quality QA-based nugget sets in three stages: (1) document-grounded \emph{nugget generation}, (2) \emph{paraphrase clustering}, and (3) \emph{nugget subselection} based on principled quality criteria. We integrate \approach nuggets with \autoargue---a recent nugget-based evaluation framework---to enable fully automatic evaluation of generated reports. We conduct extensive experiments on two cross-lingual TREC shared tasks, NeuCLIR and RAGTIME, showing strong rank correlations with both human-in-the-loop and fully manual judgments. Finally, detailed analysis of our pipeline reveals that a strong LLM nugget generator is key, and that the system rankings induced by \approach are robust to outlier systems. We facilitate future research in report evaluation by publicly releasing our code and artifacts.\footnote{\url{https://github.com/manestay/dogmatiq}}
\end{abstract}




\maketitle

\section{Introduction}
\label{sec:intro}

\begin{figure*}[ht!]
    \centering
    \includegraphics[width=\linewidth]{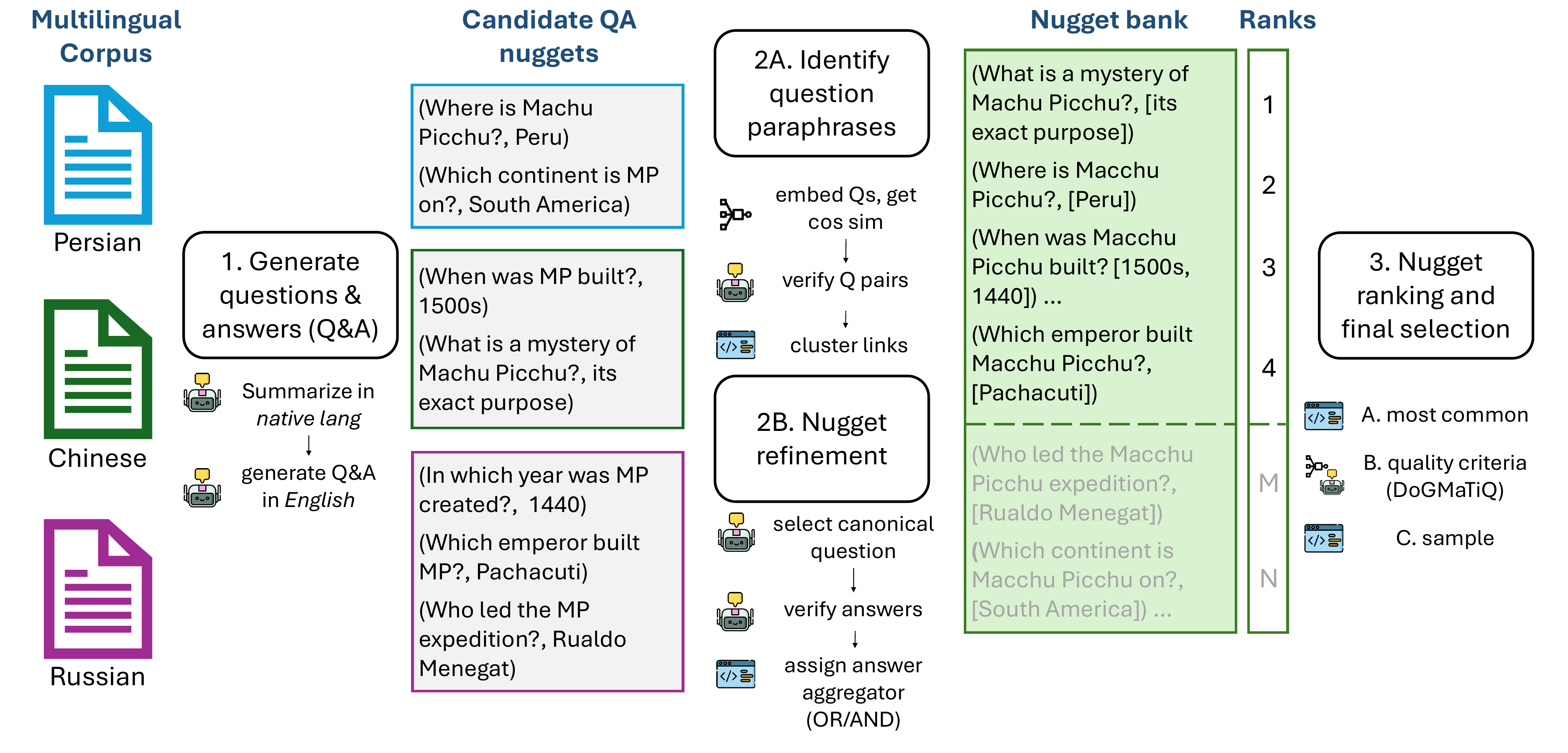}
    \Description{Illustration of the \approach\ pipeline.}
    \caption{Illustration of the \approach\ pipeline, showing the three main stages of (1) generating QA nuggets, (2) clustering nugget questions, and (3) selecting the top nuggets for inclusion in a topic's final nugget bank. Each stage consists of substeps, where icons designate an LLM, programmatic method, or ML model. QA postprocessing occurs between stages (2) and (3).}
    \label{fig:pipeline}
\end{figure*}


The recent surge of interest in retrieval-augmented generation (RAG) has created a parallel need for effective evaluation of RAG-based tasks. Of these, generation of long-form, citation-backed responses to complex queries---\emph{report generation}---is among the most challenging and has featured in numerous TREC shared tasks, including RAG~\cite{trecrag}, NeuCLIR~\cite{lawrie2025overview}, RAGTIME~\cite{ragtime}, and BioGen~\cite{gupta2024overview}.

Extending a line of research dating to the TREC 2003 Question Answering track \cite{voorhees-2003-evaluating-answers, voorhees2003overview}, recent automatic evaluation methods assess responses for how effectively they recover certain key pieces of information, dubbed \emph{nuggets}, from the underlying collection \cite{pradeep2025great, lajewska2025ginger, argue}. The majority of these methods, including AutoNuggetizer \cite{pradeep2025great} and GINGER \cite{lajewska2025ginger}, represent nuggets as short factual statements. A recall-based metric is then computed assessing the proportion of the target topic's nugget set attested by the report.

However, statement-based nuggets face limitations for contemporary reports; they display leniency biases and evaluation score compression, and struggle to scaling complex, multilingual information needs. To address this, we introduce \approach,\footnote{\underline{Do}cument-\underline{G}rounded, \underline{M}erged \underline{a}nd \underline{T}op-Cr\underline{i}teria \underline{Q}uestion-based Nuggets. \emph{Dogmatic} means adherence to a set of principles, which accurately characterizes our approach.} a nugget generation pipeline that produces nuggets in the form of \emph{question-answer (QA) pairs}. In our cross-lingual setting, this representation is essential: user information needs are expressed in English while supporting evidence may come from non-English documents. We argue QA pairs offer three critical advantages over statement-based representations. First, QA pairs decouple an information need (the question) from the content that satisfies it (the potentially diverse and multilingual answers). Second, they prevent the arbitrary overweighting of duplicate statements by consolidating redundant information into a single nugget with multiple grounded answers. Third, QA enables fine-grained assessments of nugget support via the choice of aggregation over answers---e.g.\ requiring \emph{all} answers to be provided versus requiring only a single valid answer.

\approach adapts a generate-cluster-filter paradigm for the structural complexities of nugget QA representation and report evaluation. Given a set of retrieved passages for a user query, \approach addresses QA curation in three stages: (1) document-grounded \emph{nugget generation}, (2) QA-specific \emph{paraphrase clustering} (which handles the unique semantic overlap of questions versus answers), and (3) a final \emph{nugget subselection} step driven by a learned model trained on principled linguistic and utility criteria. Critically, nugget \emph{answers} remain grounded in their original source documents, while nugget \emph{questions} anchor related answers across documents. Whereas prior approaches treat automatically generated nuggets as intermediary artifacts for automatic report generation~\cite{lajewska2025ginger, dietz2026incorporating}, our experiments treat them instead as primary units of study for automatic report evaluation.

We demonstrate the effectiveness of \approach through experiments on the report generation shared tasks from the NeuCLIR track at TREC 2024 and the RAGTIME track at TREC 2025. First, we study the viability of using \approach-generated nugget sets in lieu of human-written sets, evaluating system-level rank correlations based on nugget recall scores when using each set. Here, we find strong correlations with \approach, outperforming correlations when using nuggets generated from baseline systems and from GINGER.
Second, drawing on the recent Auto-ARGUE~\cite{autoargue} implementation of the ARGUE framework from~\citet{argue}, we evaluate whether \approach-generated nuggets can support fully automatic nugget recall scoring when combined with LLM-based nugget alignment. We compare against both organizer-provided nuggets and fully manual track-organizer scores, including manual nugget alignment.
In sum, our contributions are:
\begin{itemize}
    \item We introduce \approach, a pipeline for automatic generation of high-quality, document-grounded nuggets which enable fully automatic QA nugget-based report evaluation.
    \item We show that system rankings based on \approach-generated nuggets correlate highly with those based on human-written nuggets ($\tau=0.734$), and modestly with rankings from fully manual judgments ($\tau=0.505$).
    \item We confirm the quality of \approach-generated nuggets through several analyses, showing that \approach is robust to the influence of low-performing systems, and that the nugget recall scores it induces correlate more highly with scores from organizer-provided nuggets than with scores based on synthetic nuggets from other systems.
\end{itemize}


\section{Background}
\label{sec:background}
\textit{Nuggets} are atomic units of information and have been widely used to evaluate long-form text \cite{voorhees2003overview}. Nugget-based evaluations can broadly be split into two phases: (1) \textit{nugget creation} (for a target topic) and (2) \emph{nugget-to-text alignment}, used to determine which nuggets are attested in the target text. Here, we focus on the first phase, developing an automated pipeline for nugget generation. For the second phase, we use existing systems and compare to official judgments. 
Below, we review the two most common nugget representations as well as the literature on each phase.

\paragraph{Nugget Representations.}
Nuggets have largely been represented in one of two forms. Historically, short statements or \emph{claims} that describe key facts have been the most widely used form within IR~\cite{pradeep2025great,lin2006will,rajput2011nugget}. More recently, however, some works have focused instead on representing nuggets as QA pairs \cite{argue,sander2021exam}, which have notably also been used for summarization evaluation~\cite{deutsch2021towards,eyal2019question,nenkova2007pyramid}. The QA form has a key advantage over the claim form in that it decouples the information need (represented by the question) from the content that satisfies it (the answers), whereas a claim both implies an information need and asserts the response. This decoupling thus enables fine-grained assessments of information needs with multiple valid responses, and also enables evaluation for multilingual retrieval, where questions and answers may be in different languages.

\subsection{Nugget Creation}

\paragraph{Manual nuggets.}
Nuggets have traditionally been manually curated, and this has remained the dominant practice for benchmarking in recent years. For example, organizers of the recent TREC 2024 NeuCLIR and TREC 2025 RAGTIME report generation tasks used manually curated nuggets for official evaluation~\cite{lawrie2025overview,ragtime}. However, this curation process requires substantial human effort, as assessors must become familiar with new topics, read through documents on those topics, and distill the most important information into clear statements or QA pairs. For this reason, manual annotation does not easily scale to new topics---a significant limitation.

\paragraph{Automatic nuggets.}
On the other end of the spectrum are fully-automatic nugget generation approaches such as the Rubric Autograder Workbench \cite{farzi2024pencils,dietz2024workbench}. This system generates nuggets solely from the query and is only sometimes competitive with manually created questions \cite{farzi2024exampp}.
To remedy the lack of grounding, out approaches ground nuggets directly in retrieved documents. Other systems perform a similar grounding: The GINGER system extracts candidate statement nuggets from relevant documents, clusters them, and uses an LLM to rerank the clusters for final selection~\cite{lajewska2025ginger}. Similarly, the Crucible system generates QA nuggets, extracts source sentences that address them, and selects a top-ranked set~\cite{dietz2026incorporating}. 
%
%
At a high level, \approach shares a generate-cluster-filter paradigm with GINGER, but our framework makes three concrete, distinguishing contributions. First, compared to GINGER, each stage of \approach is explicitly tailored to cross-lingual evaluation: we use QA nuggets, cluster on questions to aggregate multilingual answers, and rank candidates with explicit SVM weighting over textual criteria rather than opaque LLM reranking. Second, both GINGER and Crucible are designed for \emph{report generation}, whereas \approach is designed as an independent framework for \emph{report evaluation}. We study circularity as discussed in Section \ref{sec:circularity}.

\subsection{Nuggets for Evaluation}
Given a nugget set for a target topic, the next phase of nugget-based evaluation is to judge whether a report attests each nugget in the set. The most intuitive approach is through nugget recall: for each nugget, determine whether it is attested by the report. This is highly compatible with the nugget statement format. Aside from reports, nuggets have been used for summarization evaluation \cite{lin2006will,nenkova2007pyramid} and ranked retrieval evaluation \cite{rajput2011nugget}.

\paragraph{ARGUE} \citet{argue} propose ARGUE, a report evaluation framework that evaluates reports along several dimensions:
\textit{completeness} in their coverage of the report request, \textit{accuracy} with their presentation of facts, and \textit{verifiability} of facts. A series of eight binary judgments are made, which are used to compute two primary evaluation metrics: nugget recall and sentence precision. The latter metric assesses whether RAG systems are able to properly cite documents for each factual statement in their reports.

\paragraph{Automating Report Evaluation}
Substantial improvements in LLM capabilities have led to their widespread use in evaluative annotation tasks, such as pairwise preference labeling \cite{macavaney2023one,arabzadeh2025benchmarking}---an evaluation paradigm often called \emph{LLM-as-a-judge}. AutoNuggetizer, one such LLM-as-a-judge framework, was introduced as part of the TREC RAG 2024 shared task for automatic judgments of reports by nugget recall~\cite{pradeep2025great}. Manual judgments were also collected, and the organizers found a $\tau=0.87$ rank correlation with the automated approach~\cite{upadhyay2024umbrela}.
As an alternative to binary nugget recall, the Rubric Autograder Workbench approach asks an LLM to rate how effectively a report addresses a specific nugget on a scale of 0--5  \cite{dietz2024workbench}. The PrefNugget system grounds the nugget generation in preference judgments of generated responses~\cite{dietz2026too}. Their work is complementary to our approach.

\autoargue is a robust, LLM-based implementation of the ARGUE evaluation framework~\cite{autoargue}. The authors show that this system's automated judgments demonstrate good system-level rank correlations with human judgments on the NeuCLIR shared task. However, because these results relied on the existence of manually-created nugget sets, they can only be considered a partially automated (\emph{human-in-the-loop}) report evaluation approach. In this work, we address this remaining manual bottleneck with \approach.

\subsection{Risks of LLM-Based Evaluation}
\label{sec:risks}
LLM-as-a-judge evaluation frameworks have achieved strong correlations with human judgments, as discussed above. However, researchers have cautioned that such evaluations are prone to certain vulnerabilities that can lead to inflated results. This is especially the case when correlations are demonstrated on only a limited set of domains and in tightly controlled settings. Our paper on automated evaluation is thus aware of, and assesses, the following risks:

\paragraph{Circular judgments} \label{sec:circularity}
It is increasingly common for the same LLM that generated a given text to also be used to score that text. This can lead to \textit{circular judgments}~\cite{clarke2024llm,dietz2026insider}, which may inflate metric scores. Critically, this circularity may not be deliberate, and may occur due to \emph{incidental} exposures to the same artifact, whether via the same underlying LLM or a similar prompt.
We study the resilience of our approach towards the Crucible vulnerability probe \cite{dietz2026insider} and find that combining explicit document grounding with the choice of a different LLM family (Claude vs Llama) defends our approach against circularity.

\paragraph{LLM ranking correlations can be inflated by low-ranked systems} \citet{clarke2024llm} note that it is most critical for evaluation systems to identify meaningful differences between \emph{top-ranked} systems, rather than across the whole distribution. When there are many under-performing systems that can easily be distinguished, the resulting rank correlations are inflated. \citet{clarke2024llm} demonstrate this by starting with a $\tau=0.89$ result between manual and automatic assessment on 75 systems, from \citet{upadhyay2024umbrela}. They progressively drop lower-ranked systems and find a decrease to $\tau=0.51$ for the top 20. We explore a related scenario in \S\ref{sec:subset} by evaluating rank correlations on a subset of high-priority systems. We find that \approach is robust to such issues, with subset rank correlations remaining nearly equivalent to those for the full set.

\section{The \approach\ Pipeline}
\label{sec:pipeline}

\approach\ consists of three main stages, each with its own substeps. The key idea is to decompose the broader task of creating an ideal nugget bank for some topic into subtasks, and address each subtask with an existing NLP method.
The goal is to generate nuggets that are both \textit{relevant} (by grounding nugget answers in relevant documents) and \textit{comprehensive} (by producing nugget questions that cover diverse facets of the user's query). \autoref{fig:pipeline} visualizes the full pipeline.
For each topic, we first generate QA nuggets for each document in some input set. Second, we identify and merge paraphrases of nugget questions. Finally, we rank the merged nuggets and select the most salient ones for inclusion in the final nugget bank. Note that, as our pipeline focuses on report evaluation, retrieval of input documents is not considered a part of \approach; as such, we use existing IR systems for this purpose.


\subsection{Stage 1: Document-Grounded QA Nugget Generation}
Generation of QA pairs from single documents is a well-explored NLP task \cite{sasazawa-etal-2019-neural, krishna-iyyer-2019-generating, lin-etal-2024-prompting, 10.24963/ijcai.2024/889}. In our case, we wish to generate English QA pairs from a multilingual document set. While earlier work addresses this through separate English generation and machine translation systems~\cite{li2023paxqa}, a single contemporary LLM suffices to handle multiple high-resource source and target languages through a summarize-then-generate approach.

\paragraph{Summarization} Directly generating QA nuggets from a document often over-indexes on document-specific details, rather than the broader information need. This was first observed by~\citet{liu2025xrag}, who used LLMs to generate QA pairs over multilingual documents. We thus follow their advice to first prompt an LLM to summarize each document, retaining only the most salient information. Summaries are generated in the document's original language to minimize loss of key information due to translation errors.

\paragraph{Generating QA Pairs} Given a summary and a user query, we prompt an LLM to generate 1--6 QA pairs, each covering a single fact that is verifiable from the document. Regardless of the summary language, we elicit the QA pairs in English.
Our prompt has clear output specifications and includes several carefully curated few-shot exemplars. Answers are generated before their questions so as to minimize hallucinations. Thus, by construction, each nugget is grounded in a specific document, and this grounding information is propagated through the rest of the pipeline. We apply the summarization and generation steps to each relevant document for a topic, obtaining a large set of candidate nuggets.

\subsection{Stage 2A: Identifying Question Paraphrases through Clustering}
Since QA pairs are generated independently for each document, the resulting candidate nuggets may have significant semantic overlap. For example, consider the following two QA pairs:

\begin{quote}
    (What is the Statue of Liberty's height?, 305 feet) \\
    (How tall is the Statue of Liberty?, 93 meters)
\end{quote}
Both the questions and answers express the same meaning, although the answers use different units. With the QA format, we can merge these two nuggets into a single nugget with a single question text, and a set of two permissible answers, along with their respective document grounding information.
In contrast, with the statement-based nuggets used in GINGER and AutoNuggetizer, these would be unnaturally treated as distinct statements.

\paragraph{Paraphrase Identification} We treat question deduplication as a paraphrase identification task. Although prompting LLMs to determine whether two sentences are paraphrases of each other is quite effective~\cite{zhou2025paraphrase}, this is expensive for large $N$ given the quadratic number of comparisons.\footnote{With only 1000 sentences, this is already $1000 \choose 2$ = 499500 comparisons.} Thus, we first identify candidate pairs with a small, fine-tuned embedding model (see \S\ref{sec:experimental}), and label pairs for which the embeddings' cosine similarity exceeds 0.9 as paraphrases.\footnote{This threshold was determined via manual experimentation on a small set of queries.} We then elicit more reliable paraphrase judgments from an LLM on this smaller candidate set. Our prompt for this task has clear instructions along with several difficult few-shot exemplars. The resulting set of verified paraphrase pairs is treated as edges in a graph, and we find larger paraphrase clusters through single-link clustering (i.e.\ connected components).

\subsection{Stage 2B: Nugget Refinement} Following clustering, we apply several further refinement steps to ensure that nuggets have high factual precision and satisfy the structural requirements of the ARGUE framework.

\paragraph{Canonical Question Selection} From each cluster of paraphrased questions, we select a single canonical question that best represents the latent information need. For this canonization step we use a zero-shot prompt which also includes the original query.

\paragraph{Answer Set Validation}
Despite our quality assurance measures in stages (1) and (2A), some generated nugget answers may still be uninformative or inaccurate. We address these issues with two layers of validation on answer sets. First, we use a regular expression to filter out uninformative responses (e.g.\ `none,' `null,' `no answer,' `unknown'). Second, we verify the factual consistency of answer sets by prompting an LLM to remove any answers it deems implausible or contradictory given the others in the set.As this involves some subjective judgment, we again ensure that our prompt includes detailed instructions and few-shot exemplars.

Finally, we remove any nugget questions with no answers remaining at the end of this process. Note that our validation can only \emph{remove} answers, thus improving the \emph{precision} of the answer sets. \emph{Recall} is much less of a concern due to the large number of candidate nuggets that still remain after this step.

\paragraph{Answer Aggregator Selection} ARGUE supports specifying a logical aggregation over nugget answers to enable accurate determination of when a nugget question is correctly addressed. In this final refinement step, we prompt an LLM to analyze the relationship among answers to each nugget question and assign that nugget an aggregator: \texttt{OR} is assigned when \emph{any} (one) answer suffices, while \texttt{AND} is assigned when \emph{all} must be provided.

\subsection{Stage 3: Final Nugget Selection}
\label{sec:our-methods}
The previous stages yield a large, comprehensive set of candidate nuggets. However, reports are almost always limited in content relative to the amount of relevant material in the collection they draw on. Thus, nugget-based evaluations must identify a privileged set of the most critical nuggets that \emph{any} high-quality report on the target topic must cover; this is the goal of stage (3). Given a ranking of nuggets, we select the top 20 for each topic, observing that human-authored nugget banks for topics in NeuCLIR and RAGTIME generally feature 10--30 nuggets.
We consider several methods for \emph{importance}-based ranking of nugget questions:

\paragraph{Method A: Most \commona} This method adopts the rationale of the pyramid method~\cite{nenkova2007pyramid}: the number of distinct documents in which a fact is mentioned can serve as a signal for the fact's importance. We thus rank nuggets by their occurrence frequency, based primarily on the number of questions in the paraphrase cluster for each nugget, breaking ties based on the total number of grounding documents across all answers.

\paragraph{Method B: Quality Criteria (\approach)}
While document frequency is a strong signal of importance, it risks overlooking nuggets that are attested in only a handful of documents but that are nonetheless essential to addressing the request.
This method takes an alternative approach, representing each nugget as a vector of 19 explicit quality criteria (detailed in \autoref{tab:quality_metrics}). These criteria, adapted from prior works~\cite{yu2025rewardanything,rosset2024researchy}, fall into two categories: \textit{linguistic quality} (e.g.\ fluency and reading level) and \textit{information utility} (e.g.\ personalization and criticality). We author LLM prompts to obtain most of these labels. We then train a Support Vector Machine (SVM) to act as a weighted average to effectively distinguish between high-quality, critical nuggets and lower-quality candidates.

\begin{table}[t]
\centering
\caption{Quality criteria used for assessing nuggets. All criteria are assessed via LLM prompting except for 1 and 2. \label{fig:quality_criteria}}
\begin{tabular}{@{}clll@{}}
\toprule
& \textbf{\#} & \textbf{Criterion} & \textbf{Scale} \\
\midrule

\multirow{3}{*}{\rotatebox{90}{Basic}}
&1 & Reading Level & 4.0--13.0 \\
&2 & Complexity & 1.0--6.0 \\
&3 & Vitality & Binary \\
\midrule

\multirow{6}{*}{\rotatebox{90}{Personalization}}
&4 & Goal Match & 0.0--1.0 \\
&5 & Background Match & 0.0--1.0 \\
&6 & Role Match & 0.0--1.0 \\
&7 & Communication Match & 0.0--1.0 \\
&8 & Scope Match & 0.0--1.0 \\
&9 & Overall & 0.0--1.0 \\
\midrule

\multirow{10}{*}{\rotatebox{90}{General}}
&10 & Fluency & 1.0--5.0 \\
&11 & Clarity & 1.0--5.0 \\
&12 & Ambiguity & 1.0--5.0 \\
&13 & Relevance & 1.0--5.0 \\
&14 & Incompleteness & 1.0--5.0 \\
&15 & Assumptiveness & 1.0--5.0 \\
&16 & Multifaceted & 1.0--5.0 \\
&17 & Knowledge Intensiveness & 1.0--5.0 \\
&18 & Subjectiveness & 1.0--5.0 \\
&19 & Reasoning Intensiveness & 1.0--5.0 \\
\bottomrule
\end{tabular}
\label{tab:quality_metrics}
\end{table}


\paragraph{Method C: \randoma} This method randomly selects 20 nuggets from the full \approach-generated candidate set, effectively ablating the subselection process.

\section{Experimental Setup}
\label{sec:experimental}
\paragraph{Datasets}
Our primary study on the efficacy of \approach focuses on the TREC 2025 RAGTIME collection~\cite{ragtime},\footnote{\url{https://trec-ragtime.github.io/index.html}} which contains news documents in English, Arabic, Chinese, and Russian. We consider the 16 test topics from the report generation task, and the 59 valid system submissions. We compare system-level correlations between rankings based on \autoargue nugget recall scores using (1) our \approach-generated nugget sets and (2) human-authored nugget sets, with the latter drawn from the official evaluation.


We study our approach on data from  TREC 2025 RAGTIME track and its predecessor the TREC 2024 NeuCLIR track~\cite{lawrie2025overview,ragtime}.
%
Both tracks are about generating English long-form reports that are grounded in news-related source documents in Chinese, Russian, Farsi or Spanish. 
Additionally, we study rank correlations between manual and automatic nugget recall judgments on TREC NeuCLIR.

Aside from rank correlations, \S\ref{sec:analysis} has experiments that study the actual scores output by different automatic nugget systems as $(x, y)$ points vs.\ manual nuggets, as well as a qualitative study on \approach nuggets for a single NeuCLIR topic.

\paragraph{Metrics}

Following ARGUE~\cite{argue}, we evaluate systems via \emph{nugget recall}: the proportion of nuggets for a given topic that are correctly addressed in a system report.\footnote{ARGUE also evaluates citation support via a \emph{sentence precision} metric, though we ignore this given our focus on nuggets.} Scores are macro-averaged across topics to produce an overall evaluation score for each system.
To assess the quality of different nugget sets, we examine their effect on nugget recall using an automatic nugget scanning system. Our experiments use \autoargue{} for determining nugget recall.\footnote{Alternatively, RUBRIC Autograder Workbench~\cite{dietz2024workbench} or AutoNuggetizer~\cite{pradeep2025great} could be used, provided they are adapted with support for the QA-nugget format.}

 Our objective is to identify nugget sets that yield a leaderboard that is maximally similar to the official leaderboard. We quantify this similarity using Spearman’s $\rho$ as well as weighted and unweighted Kendall's $\tau$, determining which nugget set best matches the official ranking. As a measure of statistical significance, we follow~\citet{autoargue} in reporting Wilcoxon paired accuracy (WPA), the probability that two Wilcoxon tests agree on the relative ranking of any given pair of runs, computed over all pairs.


\paragraph{Baseline: GINGER}
Even though GINGER \cite{lajewska2025ginger} is designed to be a RAG system, we use its nugget generation approach as an alternative system for comparison. We apply GINGER to RAGTIME topics and documents to obtain alternative nugget sets for each topic, then use \autoargue\ to score reports with GINGER-generated nuggets,\footnote{For compatibility with GINGER, in this baseline experiment we modified \autoargue's prompts to take nugget statements rather than QA nuggets.} and compute the nugget recall measure. We rank systems by this measure to obtain the leaderboard under the GINGER method. For fair comparison, the same LLM (Claude 3.5 Sonnet) backs the calls for both the GINGER baseline and \approach.

\paragraph{\approach\ Pipeline Configuration}
Below we provide implementation details for each stage of our pipeline.
\begin{description}
\item[Stage 0 (Retrieval)] We use PLAID-X~\cite{yang2024translate} as our cross-lingual IR system to retrieve input documents.

\item[Stage 1] We use Claude 3.5 Sonnet~\cite{anthropic2024claude35sonnet} to summarize the documents and generate candidate nuggets.
\item[Stage 2] For paraphrase identification (stage (2A)), we fine-tune a pretrained \href{https://huggingface.co/BAAI/bge-large-en-v1.5}{bge-large-en-v1.5} checkpoint on the Quora duplicate questions dataset.\footnote{
\url{https://huggingface.co/datasets/sentence-transformers/quora-duplicates}}
For all refinement steps (stage (2B)), we use Llama 3.3 70B Instruct~\cite{grattafiori2024llama}.
\item[Stage 3] We train an SVM classifier to select the final nuggets based on the 19 quality criteria discussed above. Positive training examples were drawn from human-written nuggets and negative examples were then mined from nuggets generated by \approach\ that were judged \emph{not} to be paraphrases of the positive examples.
\end{description}

\paragraph{Compared Systems.}
We empirically compare:
\begin{description}
    \item[Manual] Official manual evaluation scores for each system. This is our gold standard for our leaderboard correlation experiment. These scores were obtained by manually designing nuggets, then manually aligning them to responses.
    \item[\approach] Our full approach with quality criteria (method B), as described in \S\ref{sec:our-methods}. We use these nuggets with the automated evaluation system \autoargue.
    \item[\commona] A variant of our approach where we only use most common nuggets (method A).
    \item[\commonl] Same as above but using Llama for generation.
    \item[\randoma] An ablation of our approach, where we choose a subset of 20 randomly sampled nuggets (method C).
    \item[\randoml] Same as above but using Llama for generation.
    \item[Manual Nuggets] Using official manually created nuggets with \autoargue.
    \item[GINGER Nuggets] Using nuggets created by the GINGER system~\cite{lajewska2025ginger}, with \autoargue.
\end{description}





\section{Main Results}
\label{sec:evaluation}
Our main experiments show the efficacy of \approach nuggets for automated report evaluation, comparing run scores from our automated nuggets to those obtained using manually curated nuggets. For automated nuggets, we compare the three nugget selection methods from stage (3) of \approach. In \S\ref{sec:ragtime_results}, we evaluate automated nugget sets against manual nugget sets on RAGTIME, using the automated \autoargue framework to score both. In \S\ref{sec:neuclir_results}, we evaluate fully automatic scoring against fully manual reference leaderboards for both NeuCLIR and RAGTIME.

\subsection{Report Evaluation: Manual Nuggets vs.\ \approach Nuggets}
\label{sec:ragtime_results}
\begin{table}[t!]
\centering
\caption{Main results on RAGTIME report generation runs. Each row shows rank correlation metrics based on nugget recall for different nugget systems, all automated. The reference leaderboard underlying these correlations uses manual nuggets scored with \autoargue.}
\begin{tabularx}{\linewidth}{ll|XXXX}
\toprule
& Nugget System & $\rho$ & $\tau$ & Wtd. $\tau$ & WPA \\
\midrule
\multirow{4}{*}{\rotatebox{90}{Claude}}
& GINGER Nuggets & 0.793 & 0.640 & 0.594 & 0.819 \\
& \randoma & 0.737 & 0.591 & 0.639 & 0.793 \\
& \commona & 0.694 & 0.541 & 0.476 & 0.772 \\
& \approach & \textbf{0.899} & \textbf{0.734} & \textbf{0.747} & \textbf{0.866} \\
\midrule
\multirow{3}{*}{\rotatebox{90}{Llama}}
& \randoml & 0.506 & 0.366 & 0.175 & 0.681 \\
& \commonl & 0.442 & 0.296 & 0.101 & 0.648 \\
& \approachl & 0.465 & 0.319 & 0.152 & 0.656 \\
\bottomrule
\end{tabularx}
\label{tab:correlation_metrics}
\end{table}

Table~\ref{tab:correlation_metrics} shows results on RAGTIME, and we summarize our main findings below.
We first compare systems when using Claude 3.5 Sonnet for candidate nugget generation. Depending on the metric, \approach has either very high ($\rho=0.899$) or moderate correlation ($\tau=0.734$) with official leaderboards based on manually designed nuggets that are manually matched.

\randoma ($\rho=0.737$) and \commona ($\rho=0.694$) substantially underperform, even though they draw from the same candidate nuggets as \approach.
GINGER is a distant second best ($\rho=0.793, \tau=0.640$), as its nugget filtering stage is a simpler reranking. This demonstrates the effectiveness of \approach's utilization of 19 quality criteria with a learned SVM weighting. Results with Wilcoxon paired accuracy (WPA) exhibit the same trends as the other metrics.

We found that using nuggets generated by Llama is detrimental to rank correlations, with weighted $\tau$ dropping as low as 0.101.
Manual inspection confirmed these nuggets were well-formed and factually grounded, but tended to be less personalized than Claude nuggets. Circular judgments may be the culprit, as 10 systems adopted a nugget-aware approach that used Llama (\S\ref{sec:scores}). In contrast, the main \approach pipeline used Claude and avoided this.

\subsection{Report Evaluation: Fully Automatic vs.\ Fully Manual Scoring}
\label{sec:neuclir_results}
\begin{table}[t!]
\centering
\caption{Main results on NeuCLIR and RAGTIME report generation runs. Each row shows rank correlation metrics based on nugget recall for different nugget systems (manual or automated). The reference leaderboards underlying the correlations use manual nuggets scored by human assessors.}
\begin{tabularx}{\linewidth}{ll|XXXX}
\toprule
& Nugget system & $\rho$ & $\tau$ & Wtd. $\tau$ & WPA \\
\midrule
\rowcolor{gray!15} \cellcolor{white} & \randoma & 0.720 & 0.552 & 0.571 & 0.775 \\
\rowcolor{gray!15} \cellcolor{white} & \approach & \textbf{0.768} & \textbf{0.586} & \textbf{0.614} & \textbf{0.792} \\ \cmidrule{2-6}
\rowcolor{gray!15} \multirow{-3.3}{*}{\cellcolor{white}\rotatebox{90}{NeuCLIR}} & Manual Nuggets & {0.804} & {0.630} & {0.640} & {0.815} \\
\midrule
\rowcolor{gray!15} \cellcolor{white} & GINGER Nuggets & 0.579 & 0.421 & 0.302 & 0.708 \\
\rowcolor{gray!15} \cellcolor{white} & \approachl &  0.653 & 0.505 & 0.287 & \textbf{0.751} \\
\rowcolor{gray!15} \cellcolor{white} & \randoma & 0.507 & 0.372 & 0.291 & 0.686 \\
\rowcolor{gray!15} \cellcolor{white} & \approach & \textbf{0.664} & \textbf{0.505} & \textbf{0.383} & 0.750 \\ \cmidrule{2-6}
\rowcolor{gray!15} \multirow{-5.3}{*}{\cellcolor{white}\rotatebox{90}{RAGTIME}} & Manual Nuggets & 0.748 & 0.578 & 0.589 & 0.786 \\
\bottomrule
\end{tabularx}
\label{tab:corr_manual}
\end{table}

We next assess how closely \approach{} can approximate fully manual evaluations on both NeuCLIR and RAGTIME, where both the nuggets and relevance assessments are produced by humans. Our fully automated evaluation uses \approach{} to generate nuggets and then applies the automated nugget alignment method from \autoargue{} to obtain relevance labels.

We compare this setting against the Manual Nuggets baseline, a semi-automatic approach that uses human-curated nuggets but applies the same automated nugget alignment from \autoargue{}. Since this baseline removes the uncertainty introduced by automatic nugget generation, it serves as an upper bound on the correlations we can expect from our fully automated setting.%
\footnote{The ``Manual Nuggets'' setting for NeuCLIR is a reproduction of an experiment from \cite{autoargue}. While they report $\tau=0.73$, we find $\tau=0.63$. After discussion with those authors, we determined this is due to LLM nondeterminism, as well as revised, stricter versions of \autoargue. This section's results regardless are self-contained so are still valid.}

Table~\ref{tab:corr_manual} details the rank correlation results for both datasets. On NeuCLIR, \approach achieves strong correlations to the Manual Nuggets upper bound. For Kendall's $\tau$, \approach recovers 93\% of the maximum possible correlation ($\tau=0.586$ vs.\ $0.630$), while for Spearman's $\rho$, it recovers 95\% of it ($\rho=0.768$ vs.\ $0.804$). Across both metrics, it consistently outperforms the \randoma baseline (e.g., $\tau=0.586$ vs.\ $0.552$). This demonstrates that fully automatic evaluation using \approach nuggets is a highly viable substitute that significantly alleviates assessor burden.

On RAGTIME,\footnote{Our Manual Nugget $\tau=0.578$ on RAGTIME closely aligns to the $\tau=0.549$ reported in Table 6 of the RAGTIME'25 paper \cite{ragtime}.} \approach is again the best-performing fully automated system across most metrics, achieving $\tau=0.505$ and $\rho=0.664$. This represents only a 13\% drop in Kendall's $\tau$ relative to the Manual Nuggets baseline ($\tau=0.578$). Furthermore, \approach handily outperforms the alternative automated pipelines, beating \randoma ($\tau=0.372$) and GINGER ($\tau=0.421$) by wide margins. While \approachl achieves a comparable unweighted $\tau$ ($0.505$), it suffers a substantial drop in weighted $\tau$ ($0.287$ vs.\ $0.383$ for \approach), indicating that Claude-generated nuggets are markedly better at accurately ordering the highest-ranked systems.

\section{Analysis}
\label{sec:analysis}
\subsection{Robustness to Underperforming Systems}
\label{sec:subset}

\begin{table}[t!]
\centering
\setlength{\tabcolsep}{4pt}
\caption{Results on a subset of RAGTIME systems marked as ``highest priority'' for each team (12 total). Each system is evaluated with \autoargue. Each cell also shows the difference $\delta$ between the metric value on these 12 runs and the same metric value on all 59 runs. The first column indicates whether nuggets were generated with Llama or Claude.}
\begin{tabularx}{\linewidth}{ll|XXXX}
\toprule
& Nuggets & $\rho$ & $\tau$ & Wtd. $\tau$ & WPA \\
\midrule

\multirow{6}{*}{\rotatebox{90}{Claude}} &
GINGER Nuggets & 0.657 \newline ($\delta$ -0.14) & 0.545 \newline ($\delta$ -0.10) & 0.528 \newline ($\delta$ -0.07) & 0.773 \newline ($\delta$ -0.05) \\
& \randoma & 0.796 \newline ($\delta$ +0.06) & 0.677 \newline ($\delta$ +0.09) & 0.713 \newline ($\delta$ +0.07) & 0.833 \newline ($\delta$ +0.04) \\
& \approach & 0.883 \newline ($\delta$ -0.02) & 0.748 \newline ($\delta$ +0.02) & 0.687 \newline ($\delta$ -0.06) & 0.864 \newline ($\delta$ -0.00) \\ \midrule
\multirow{4}{*}{\rotatebox{90}{Llama}}
& \randoml & 0.608 \newline ($\delta$ +0.10) & 0.515 \newline ($\delta$ +0.15) & 0.455 \newline ($\delta$ +0.28) & 0.758 \newline ($\delta$ +0.08) \\
& \approachl & 0.580 \newline ($\delta$ +0.12) & 0.485 \newline ($\delta$ +0.17) & 0.373 \newline ($\delta$ +0.22) & 0.742 \newline ($\delta$ +0.09) \\
\bottomrule
\end{tabularx}
\label{tab:correlation_metrics_subset}
\end{table}
This experiment addresses the risk of rank correlations being inflated by many low-performing systems (see \S\ref{sec:risks}). Note that while \citet{clarke2024llm} evaluate Kendall's $\tau$ strictly among the top-$k$ systems of the official leaderboard, we simulate a different, real-world scenario: one in which we want to use \approach to rank a subset of likely strong systems without prior knowledge of the outcome of a manual assessment. As we cannot know which systems perform best, we use as a proxy the participants' own priority labels on runs, taking the highest priority run per team. This yields a 12-run subset on which we perform rank correlation analysis.

Table~\ref{tab:correlation_metrics_subset} presents the results. We find that the metrics for \approach are stable across the board, with much smaller relative changes compared to all other systems ($\tau=+0.01$ or 2\% relative to $\tau$ computed over all runs). In contrast, GINGER suffers a $\tau=-0.1$/14.8\% relative drop.

Interestingly, all other variants of our approach see relatively large improvements in correlation over the subset compared to the full set of runs. This could be due to teams submitting multiple very similar systems, and to the weaker variants of \approach thus struggling to discriminate between them. That the full version of \approach does not suffer from this issue illustrates its robustness to variation in the set of systems under evaluation.


\subsection{Raw Nugget Recall Scores}
\label{sec:scores}
\begin{figure}[t!]
    \centering
    \includegraphics[width=.7\linewidth]{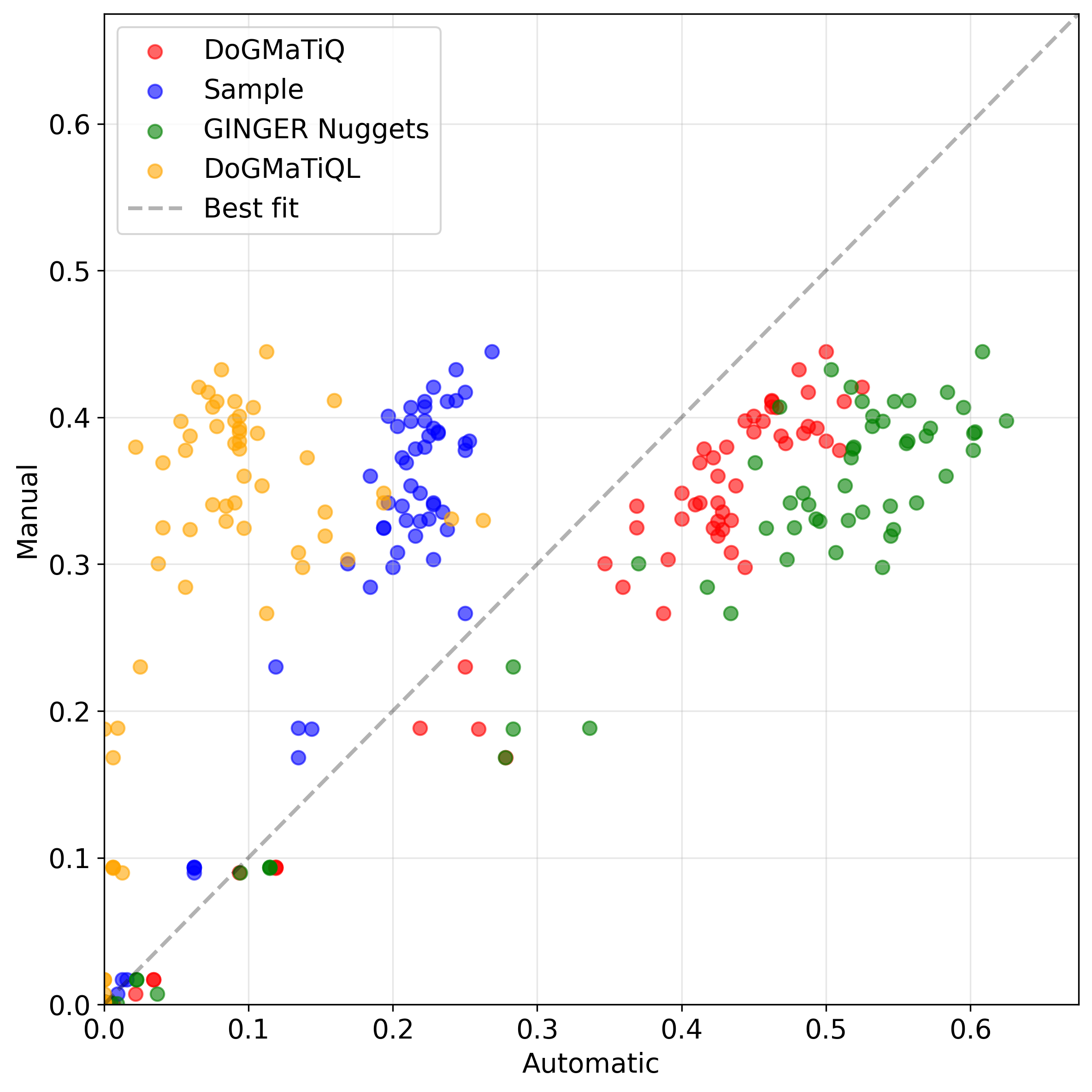}
    \Description{Scatter plot for automated nugget sets scores.}
   \caption{Scatter plot comparing macro-average nugget recall using manual nuggets vs.\ automated nugget sets for RAGTIME runs, scored with \autoargue. Above-diagonal points denote systems whose manual score is higher than under \approach, while below-diagonal points denote systems where the reverse is true. The cluster of yellow points (left) use the same LLM as the Crucible system does (Llama), and thus is a subversion probe, studying effects of circularity.}
    \label{fig:macro_systemscatter}
\end{figure}

Inspection of raw nugget recall scores under different evaluation methods reveals additional insights. This experiment follows the RAGTIME setting with \autoargue judgments from \S\ref{sec:ragtime_results}. \autoref{fig:macro_systemscatter} shows a scatter plot where the $x$-coordinate of each point gives the nugget recall score using automatically generated nuggets from a given system and where the $y$-coordinate gives the nugget recall score using manually created nuggets.

The points for \approach{} lie closest to the diagonal, with a reasonably narrow spread and no clear outliers. By contrast, GINGER Nuggets shows a wider spread and consistently assigns higher scores than the manual nugget evaluation. This suggests leniency bias, which can make it harder to distinguish strong systems from the best-performing ones. In contrast, \randoma consistently yields scores lower than those from manually generated nuggets, with a narrow spread and a few outliers. Finally, \approachl shows the lowest scores and the widest spread.

\autoref{fig:macro_systemscatter} thus illustrates that \approach produces nugget sets whose scores most closely match those found under evaluation with human-written nuggets. GINGER instead tends to produce more easily answerable nuggets, while \randoma and \approachl tend to produce more difficult ones.

\paragraph{Study on Circularity} \autoref{fig:macro_systemscatter} also offers insights into circular judgments (\S\ref{sec:risks}). Ten systems used \crucible, a nugget-aware report generation system~\cite{dietz2026incorporating}. Considering the yellow points for \approachl in \autoref{fig:macro_systemscatter}, we observe a distinct cluster of relative outliers shifted to the right of the others. These outlier systems receive highly inflated automated scores under \approachl, which strongly contributes to the decreased rank correlations. Further evidence comes from \approachl scores being much higher on the top-system subset than on all systems (\S\ref{sec:subset}).

In contrast, \approach avoids this circularity by relying on explicitly grounded nuggets and using a different LLM family (Claude). If we were to run \crucible with Claude, we would likely observe another outlier cloud. The observed performance drop for \approachl is therefore primarily a symptom of evaluation circularity inherent to the RAGTIME track composition, rather than a structural flaw in our pipeline. These results thus emphasize the importance of explicitly grounding nuggets and decoupling the evaluation model from generation models~\cite{clarke2024llm,dietz2026insider}.

\subsection{System-Level vs.\ Topic-Level Scores}
\begin{figure*}[t!]
    \centering
    \begin{minipage}{0.37\linewidth}
        \centering
        \includegraphics[width=\linewidth]{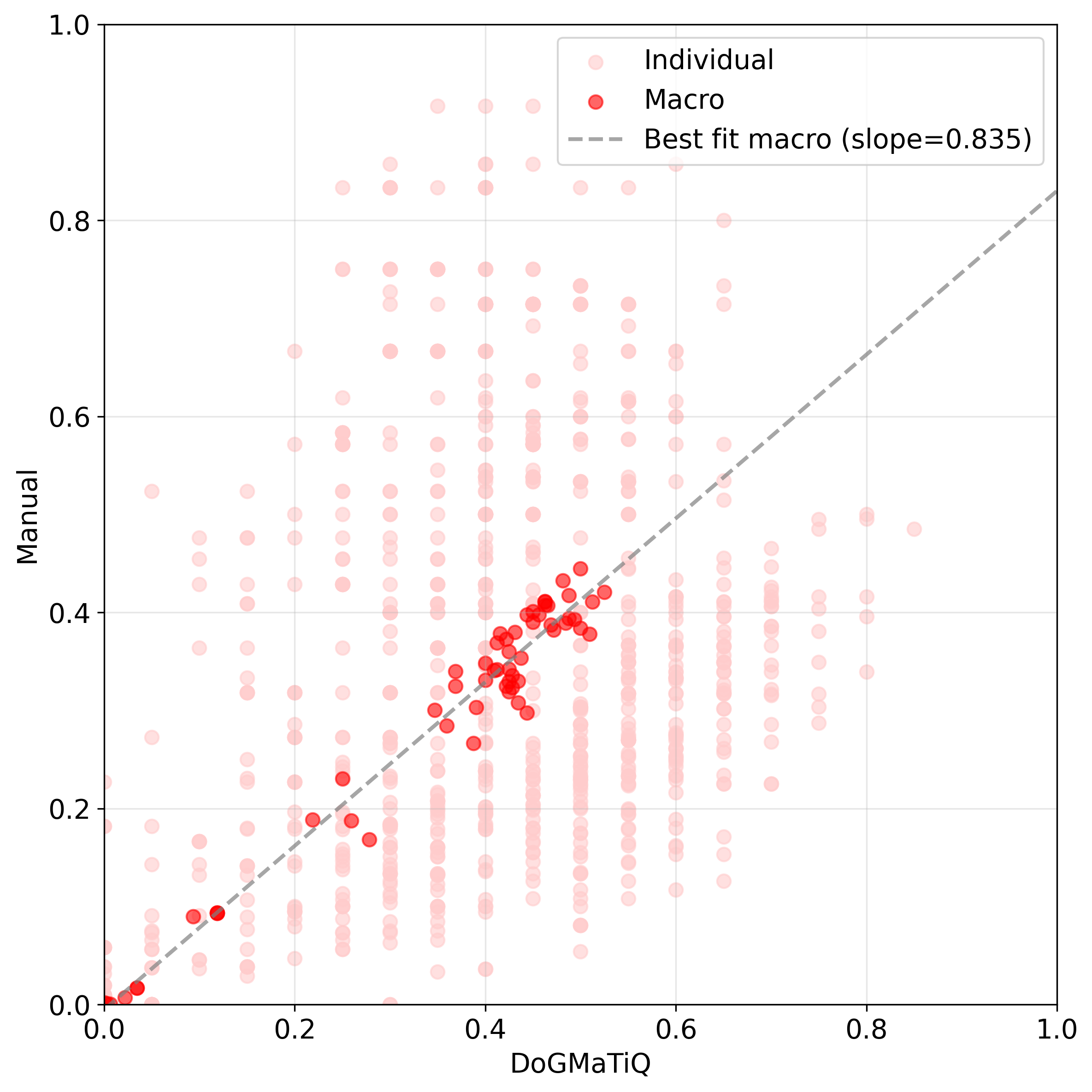}
    \end{minipage}
    \begin{minipage}{0.37\linewidth}
        \centering
        \includegraphics[width=\linewidth]{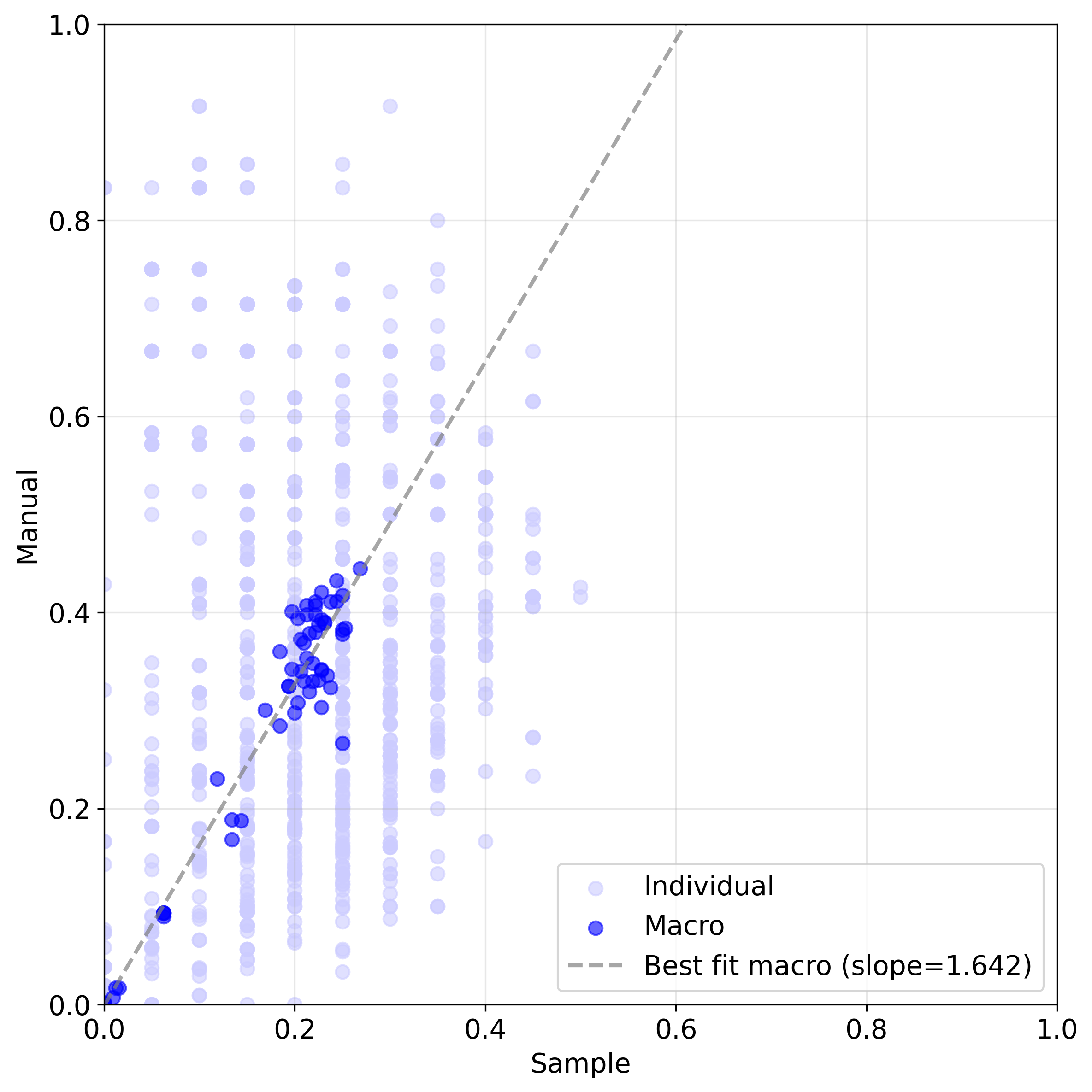}
    \end{minipage}
    \begin{minipage}{0.37\linewidth}
        \centering
        \includegraphics[width=\linewidth]{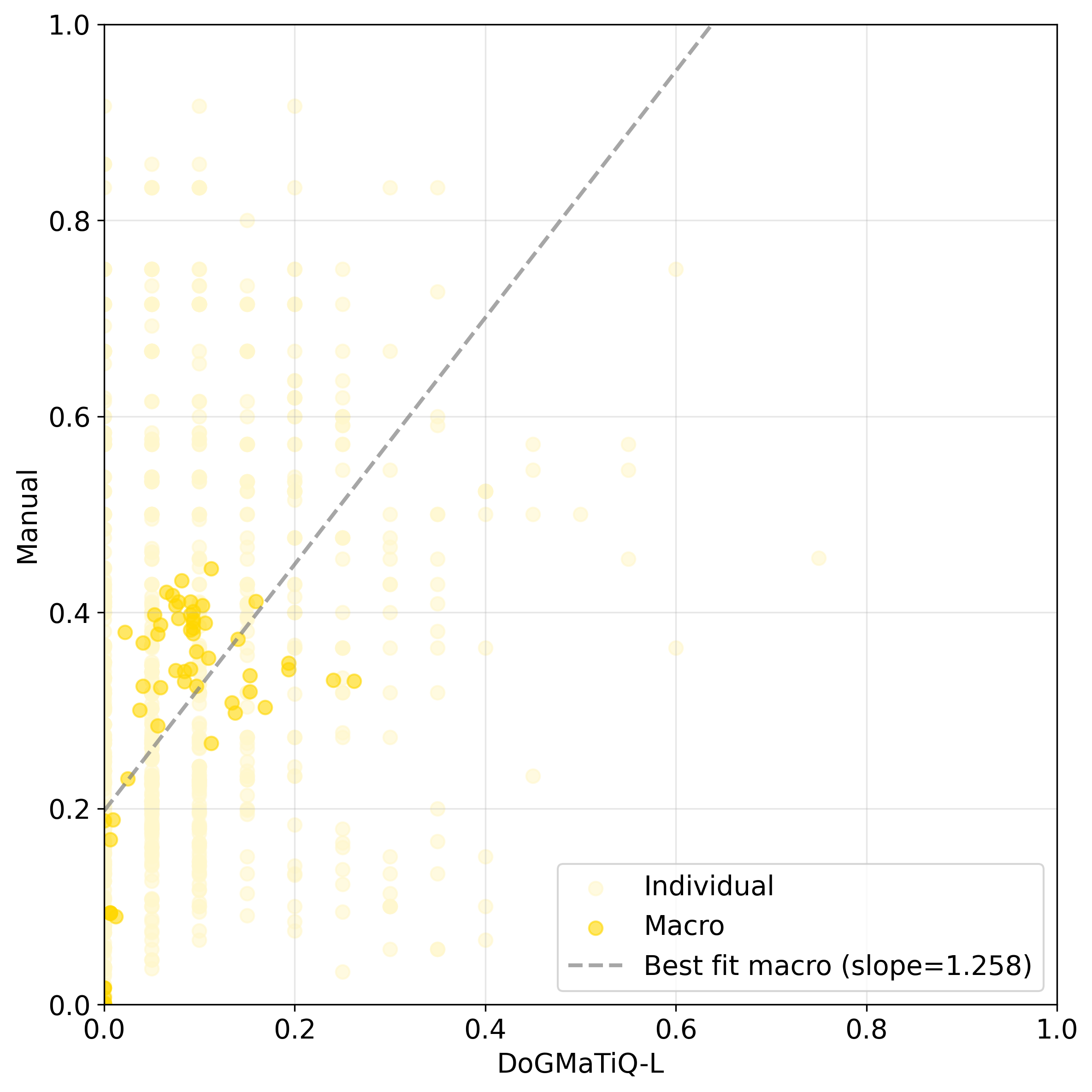}
    \end{minipage}
    \begin{minipage}{0.37\linewidth}
        \centering
        \includegraphics[width=\linewidth]{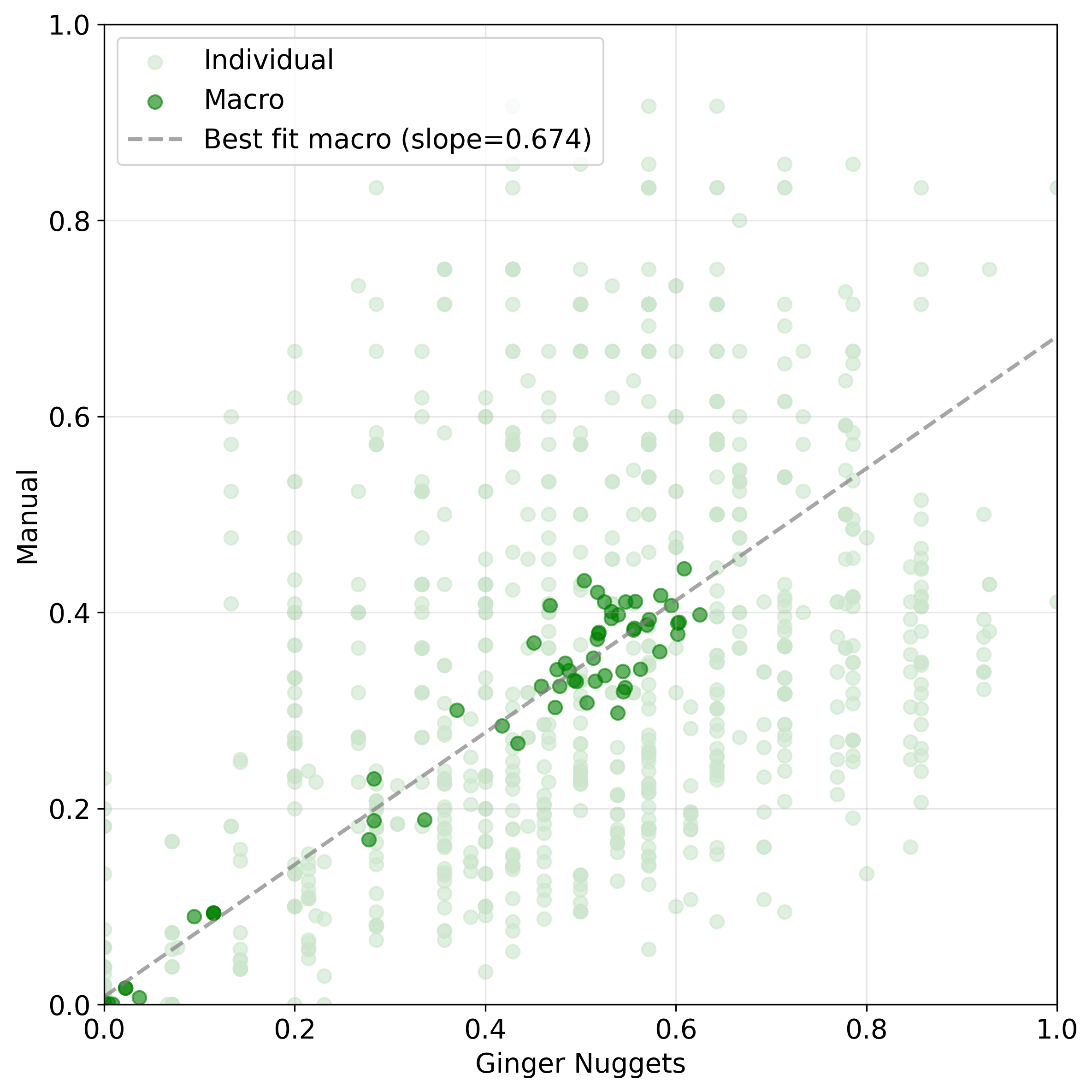}
    \end{minipage}
    
    \Description{Scatter plots for topic and macro-level scores, 4 nugget systems.}
    \caption{Scatter plots comparing nugget recall using manual nuggets vs.\ various automated nuggets for RAGTIME runs, scored with \autoargue. Darker dots are systems' macro-average scores over topics, while lighter dots plot systems' per-topic scores.}
    \label{fig:scatter plots_all}
\end{figure*}

So far, we have considered only system-level scores, which represent macro-averages across topics. We now turn to per-topic scores: \autoref{fig:scatter plots_all} presents scatter plots analogous to the one in \autoref{fig:macro_systemscatter}, but where points represent nugget recall scores for individual runs on individual topics using various auto-nugget sets. 

Focusing our analysis on \approach (top left), we corroborate findings from \cite{upadhyay2024large} that despite strong system-level rank correlations, per-topic points exhibit high variance. Such variance is unsurprising: Creating nugget sets is an inherently subjective process, and different annotators---whether human or machine---will differ in which pieces of information they deem most central to a given topic~\cite{lawrie2025overview}. Running a system over multiple topics can be viewed analogously to drawing samples from a larger population of valid nuggets. \citet{upadhyay2024large} estimate that having scores on 10 topics suffices to achieve good rank correlations, which is consistent with our results using \approach on these 16 RAGTIME topics.

As for nugget recall for other automated nuggets, we see similarly high variance for per-topic points. The best-fit line for \randoma and \approachl have steeper slopes than \approach, indicating that the QA nuggets taken as a set tend to be harder to answer than humans' would write. The inverse is true for GINGER nuggets, which are easier to answer.

\begin{figure}[t!]
    \centering
    \begin{minipage}{0.48\linewidth}
        \centering
        \includegraphics[width=\linewidth]{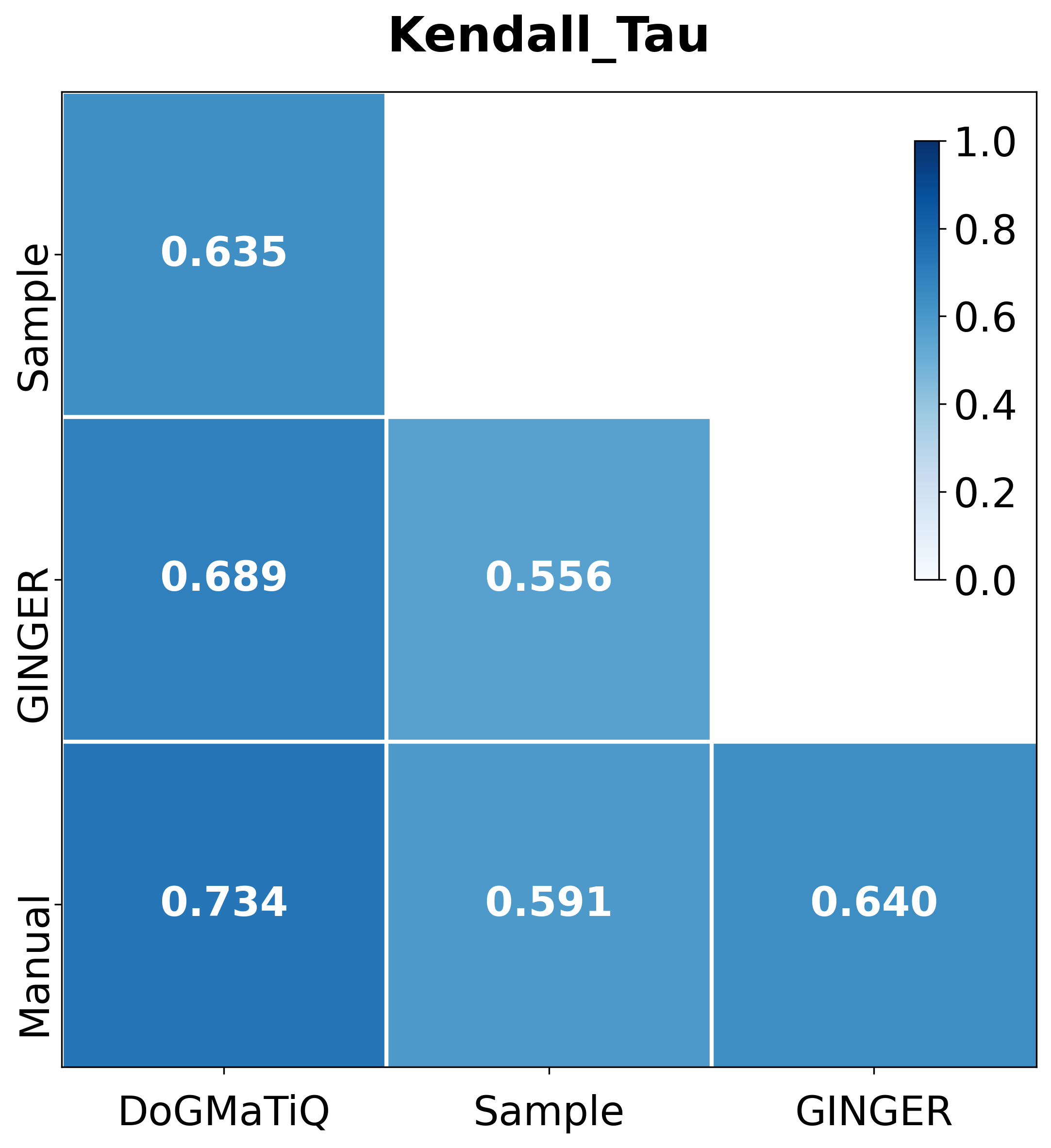}
    \end{minipage}\hfill
    \begin{minipage}{0.48\linewidth}
        \centering
        \includegraphics[width=\linewidth]{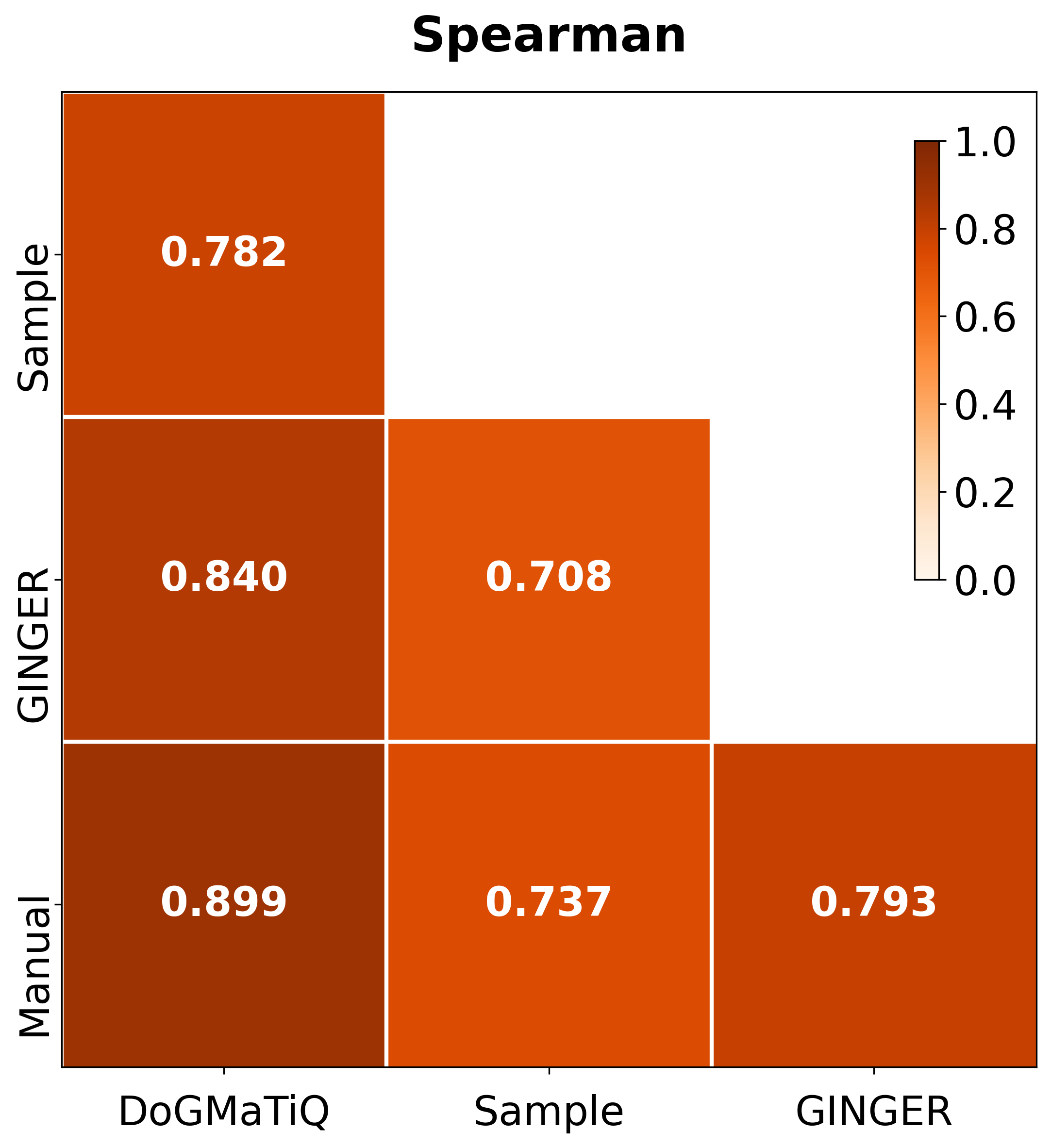}
    \end{minipage}

    \Description{Side-by-side heatmaps of rank correlation between nugget sets.}
    \caption{Heatmaps showing rank correlation with official fully manual leaderboards ($\tau$ and $\rho$) when scoring RAGTIME topics with different nugget sets. We compare three automated systems with official leaderboards, yielding nine unique comparisons. The bottom row numbers are also provided in \autoref{tab:correlation_metrics}. This shows that \approach obtains the best correlation with manual leaderboards.}
    \label{fig:heatmap_rank_corr}
\end{figure}

\subsection{Rank Correlations Between Nugget Sets}


Next, we consider how the nugget sets generated by different automatic systems compare to one another. \autoref{fig:heatmap_rank_corr} shows system-level rank correlations between nugget recall scores on RAGTIME runs under nugget sets generated by different methods. First, we note that scores using \approach nuggets correlate better with those based on manual nuggets ($\rho=0.899, \tau=0.734$) than with those based on any automatic system---once again testifying to \approach's effectiveness. Second, we find that \approach and GINGER rankings correlate more strongly with each other than either does with \randoma---emphasizing that these systems' nugget selection step is important to their success. Third, we observe that correlations among all four automated systems largely fall in the moderate range. This could be due in part to limitations of the LLM judge from \autoargue used here to assess nugget recall, and it is possible we would observe higher correlations under human assessments of nugget correctness using the same nugget sets.



\begin{figure*}[t]
  \centering
  \small
  \begin{minipage}{0.98\textwidth}
\begin{tcblisting}{mylisting}
Clearly matched:  2
1)  GEN: What are the ancient particles in the Murchison meteorite called?
   GOLD: The presolar grains in the Murchison meteorite were mostly composed of what?  SIM: 0.732
2)  GEN: Where did the Murchison meteorite fall in 1969?
   GOLD: Where did the Murchison meteorite make landfall?  SIM: 0.745
Unclearly matched: 8
3)  GEN: How many pre-solar grains were analyzed by the research team?
   GOLD: How does the age of the grains compare to that of our Solar System?  SIM: 0.643
4)  GEN: How many ancient dust particles were extracted from the Murchison meteorite?
   GOLD: How heavy was the Murchison meteorite?  SIM: 0.634
5)  GEN: How old is 60% of the stardust in the Murchison meteorite?
   GOLD: How old were these grains?  SIM: 0.521
6)  GEN: Which university's researchers discovered asymmetric molecules in meteorites?
   GOLD: What advances could the discovery of water molecules in the Murchison meteorite lead to?  SIM: 0.693
7)  GEN: What ancient materials were found in the Murchison meteorite?
   GOLD: What did the scientists use pieces of the meteorite for?  SIM: 0.608
8)  GEN: What method was used to discover sugars in the Murchison meteorite?
   GOLD: What does the discovery of sugars in the meteorite indicate about the origin of life on Earth?  SIM: 0.874
9)  GEN: What sugars were found in the Murchison meteorite?
   GOLD: What if any organic compounds were found in the Murchison meteorite?  SIM: 0.691
10)  GEN: How long before the solar system was the oldest material in the Murchison meteorite formed?
   GOLD: What other materials do we have that are older than the Solar System?  SIM: 0.605
\end{tcblisting}
\Description{Comparison of nuggets from \approach\ and gold banks for NeuCLIR topic 361.}
\captionof{figure}{Comparison of nuggets from \approach (GEN) and human-written (GOLD) for NeuCLIR topic 361. SIM is the pairwise cosine similarity from a paraphrase detection model. Each GOLD nugget is assigned to its closest GEN nugget through stable matching. ``Clearly'' and ``Unclearly'' matched pairs are judged manually by the authors.}\label{fig:case2}
\end{minipage}
\end{figure*}

\subsection{Qualitative Analysis}
Lastly, we present qualitative analysis comparing the set of 10 nuggets created by human assessors against the set generated by \approach for topic 361 (on the ``Murchison meteorite, its discovery and composition'') from NeuCLIR. We use our paraphrase detection model (\S\ref{sec:experimental}) and a stable matching algorithm to assign each manually written nugget to its most similar nugget from the set generated by \approach.

\autoref{fig:case2} presents the resulting alignment. The matched pairs in (1) and (2) are effectively synonymous; the questions differ slightly in their level of detail, but they seek the same information and feature the same answers (not pictured). Although the remaining eight pairs in the alignment are not synonymous, they nonetheless cover many of the same topics (e.g.\ the composition of the meteorite, the age of the materials it contains, and scientific analysis conducted on it). Furthermore, the level of specificity and the style are not noticeably different between the two sets. Briefly highlighting several pairs, we note that for pair (5), the GOLD nugget is under-specified and requires additional context, while the GEN nugget is specific and standalone. The questions in pair (6) are entirely distinct, though the GOLD nugget is more narrowly tailored to the topic. Finally, for pair (9), although the questions are similar and the answers are the same, the GEN nugget is more specific in the types of compounds it asks about (sugars vs.\ organic compounds).


\section{Conclusion}
We have introduced \approach, a pipeline for generating high-quality, document-grounded QA nuggets. Prior work in automated report evaluation has focused on simple, statement-based nuggets. In contrast, \approach produces QA nuggets, which allow for a much richer representation of information needs versus the information that satisfies them, and are compatible with the ARGUE framework for report evaluation. We showed that using \approach nugget sets for automated report evaluation achieves strong correlation scores with human-written nugget sets.

We further investigated the design decisions for each stage of the \approach pipeline, finding that success relies heavily on combining a strong LLM for nugget generation, and weighted quality criteria for nugget selection. Each stage of our pipeline mitigates errors and refines precision from earlier stages. Our further analyses show that \approach system rankings 
yield macro-level scores that correspond much better to those from human-written nuggets. By explicitly grounding nuggets in the source documents and utilizing a different LLM family, our approach can effectively ward against circularity.


We emphasize that \approach and related methods are meant to complement, rather than replace, expert human effort. Automated methods enable fast, scalable evaluations for arbitrary generated reports, helping researchers understand the topics and user profiles where current systems underperform.
As generative AI continues to reshape how information is created and consumed, it has never been easier to use these systems for generative purposes, making the challenge of evaluating quality ever more urgent. With \approach, we take a step toward closing that gap: we provide automated tools to assess the degree to which different information systems can satisfy information needs.



\begin{acks}
This research project was pursued as a part of the SCALE 2025 program at the HLTCOE at Johns Hopkins University. We thank the fellow SCALE participants for their discussions and feedback throughout the course of this project, including Hannah Recknor, Rebecca Kotula, Jia-Huei Ju, and Dayeon Ki.
\end{acks}

\section*{Ethics and Privacy Statement}
We acknowledge that utilizing AI to provide users with information, and specifically for our work, to generate artifacts (nuggets) for evaluation of such information, is an area of ethical consideration. The fact is that AI-generated text is all over the internet, and is prominently displayed in search results, and directly accessed by end-users through AI assistant websites. Our work, then, assists in evaluation of this generated long-form report text. We have carefully written our paper to avoid over-claiming the capabilities of AI evaluation and generation; we only report that the correlations with manual nuggets and manual evaluations are reasonable.

\bibliographystyle{ACM-Reference-Format}
\bibliography{biblio}

\end{document}